\documentclass[11pt,a4paper]{article}

\usepackage[T1]{fontenc}
\usepackage[utf8]{inputenc}
\usepackage{authblk}
\usepackage{pdfsync}
\usepackage{graphicx}
\usepackage{amssymb,amsfonts,amsmath,dsfont,amsthm,thmtools,mathtools}
\usepackage{verbatim}
\usepackage{varwidth}
\usepackage{algorithm,algpseudocode,lipsum}
\usepackage{caption}
\usepackage{framed}
\usepackage{xcolor}
\usepackage{lipsum}
\usepackage{anysize}

\newtheorem{remark}{Remark}

\usepackage{kpfonts}

\newcommand{\eni}{\begin{equation}}
\newcommand{\enf}{\end{equation}}
\newcommand{\be}{\begin{equation}}
\newcommand{\ee}{\end{equation}}

\newcommand{\br}{\begin{rem}}
\newcommand{\er}{\end{rem}}
\newcommand{\bex}{\begin{ex}}
\newcommand{\eex}{\end{ex}}

\newcommand{\bt}{\begin{thm}}
\newcommand{\et}{\end{thm}}
\newcommand{\bp}{\begin{prop}}
\newcommand{\ep}{\end{prop}}
\newcommand{\bq}{\begin{prob}}
\newcommand{\eq}{\end{prob}}
\newcommand{\bd}{\begin{defi}}
\newcommand{\ed}{\end{defi}}

\newcommand{\bc}{\begin{center}}
\newcommand{\ec}{\end{center}}
\newcommand{\bn}{\begin{enumerate}}
\newcommand{\en}{\end{enumerate}}
\newcommand{\bi}{\begin{itemize}}
\newcommand{\ei}{\end{itemize}}
\newcommand{\mfi}{\begin{eqnarray*}}
\newcommand{\mff}{\end{eqnarray*}}
\newcommand{\mfni}{\begin{eqnarray}}
\newcommand{\mfnf}{\end{eqnarray}}

\newtheorem{thm}{Theorem}
\newtheorem{defi}{Definition}

\newtheorem{prop}{Proposition}
\newtheorem{rem}{Remark}
\newtheorem{ex}{Example}

\newtheorem{prob}{Problem}

\providecommand{\nor}[1]{\left\lVert {#1} \right\rVert}

\providecommand{\scal}[2]{\left\langle{#1},{#2}\right\rangle}

\newcommand{\R}{\mathbb R}

\newcommand{\TT}{\mathcal T}

\newcommand{\G}{\mathcal G}

\begin{document}

\title{Deep Convolutional Networks are
Hierarchical Kernel Machines}

\author[1,2]{Fabio Anselmi}
\author[1,2,3]{Lorenzo Rosasco}
\author[4]{Cheston Tan}
\author[1,2,4]{Tomaso Poggio}
\affil[1]{Center for Brains Minds and Machines, Massachusetts Institute of
Technology, Cambridge, MA 02139.}
\affil[2]{Laboratory for Computational Learning, Istituto Italiano di Tecnologia and Massachusetts Institute of
Technology.}
\affil[3]{DIBRIS, Universit\'a degli studi di Genova, Italy, 16146.}
\affil[4]{Institute for Infocomm Research, Singapore, 138632.}

\maketitle
\let\thefootnote\relax\footnote{Email addresses: anselmi@mit.edu;
  lrosasco@mit.edu; cheston-tan@i2r.a-star.edu.sg; corresponding
  author: tp@ai.mit.edu. The main part of this work was done at the Institute for Infocomm Research with
  funding from REVIVE}

\begin{abstract}
  In i-theory a typical layer of a hierarchical architecture consists
  of HW modules pooling the dot products of the inputs to the layer
  with the transformations of a few templates under a group. Such
  layers include as special cases the convolutional layers of Deep
  Convolutional Networks (DCNs) as well as the non-convolutional
  layers (when the group contains only the identity).  Rectifying
  nonlinearities -- which are used by present-day DCNs -- are one of
  the several nonlinearities admitted by i-theory for the HW
  module. We discuss here the equivalence between group averages of
  linear combinations of rectifying nonlinearities and an associated
  kernel.  This property implies that present-day DCNs can be exactly
  equivalent to a hierarchy of kernel machines with pooling and
  non-pooling layers. Finally, we describe a conjecture for
  theoretically understanding hierarchies of such modules. A main
  consequence of the conjecture is that hierarchies of trained HW
  modules minimize memory requirements while computing a selective and
  invariant representation.
\end{abstract}

\section{Introduction}
The architectures now called Deep Learning Convolutional networks
appeared with the name of convolutional neural networks in the 1990s
-- though the supervised optimization techniques used for training
them have changed somewhat in the meantime. Such architectures have a
history that goes back to the original Hubel and Wiesel proposal of a
hierarchical architecture for the visual ventral cortex iterating in
different layers the motif of simple and complex cells in V1. This
idea led to a series of quantitative, convolutional cortical models
from Fukushima (\cite{Fukushima1980}) to HMAX (Riesenhuber and
Poggio, \cite{riesenhuber2000}). In later versions (Serre et al., \cite{serre2007})
such models of primate visual cortex have achieved object recognition
performance at the level of rapid human categorization. More recently,
deep learning convolutional networks
%(or generalizations like SimNets \cite{CohenS14})
 trained with very large labeled
datasets (Russakovsky et al. \cite{russakovsky2014}, Google \cite{LastConvnets2014}, Zeiler and Fergus
 \cite{Zeiler2013visualizing}) have achieved impressive
performance in vision and speech classification tasks. The performance
of these systems is ironically matched by our present ignorance of why
they work as well as they do. Models are not enough. A theory is required for a satisfactory
explanation and for showing the way towards further progress. This
brief note outlines a framework towards the goal of a full
theory.

Its organization is as follows. We first discuss how i-theory applies
to existing DLCNs.  We then show that linear combinations of
rectification stages can be equivalent to kernels. Deep Learning
Convolutional Networks can be similar to hierarchies of
HBFs (\cite{Poggio89atheory}).

%%%%%%%%%%%%%%%%%%%%%%%%%%%%%%%%%%%%%%%
\section{DCNs are hierarchies of kernel machines}
In this section, we review the basic computational units
composing deep learning architectures of the convolution type.  Then
we establish some of their mathematical properties by using i-theory as described in
\cite{anselmi2013unsupervised,Anselmi2015,Rosascokernels2015}.

\subsection{DCNs and i-theory}
%%%%%%%%%%%%%%%%%%%%%%%%%%%%%%%%%%
The class of learning algorithms called deep learning, and in
particular convolutional networks, are based on a basic
operation in multiple layers. We describe it
using the notation of i-theory.

The operation is the inner product of an input with another
point called a template (or a filter, or a kernel), followed by a non
linearity, followed by a group average. The output of the first two steps can be seen to roughly
correspond to the neural response of a so called simple cell
\cite{Hubel1962}. The collection of inner products of a given input
with a template and its {\em transformations} in i-theory corresponds to a so called
convolutional layer in DCNs.  More precisely, given a template $t$ and its
transformations $gt$, here $g\in \G$ is a finite set of
transformations (in DCNs the only transformations presently used are
translations), we have that each input $x$ is mapped to $\scal{x}{gt},
\quad g\in \G$.  The values are hence processed via a non linear
{\em activation} function, e.g. a sigmoid
$(1+e^{-s})^{-1}$, or a rectifier $|s+b|_+=\max\{-b,s\}$ for $s,b\in
\R$ (the rectifier nonlinearity was called ramp by Breiman\cite{Breiman1991}). In summary, the first operation unit can be described, for
example by
$$
x\mapsto |\scal{x}{gt}+b|_+.
$$

%Note that, in the case of translations, for all $s,\tau\in \R$,   $g x(s)=g_\tau x(x)=x(s+\tau)$, so that $\scal{x}{g^{-1}t}=x\star t(\tau)$ provided  $t$ is symmetric.

The last step, often called {\em pooling},  aggregates in a single output the values of the different inner products previously computed
that  correspond to transformations of the same template,  for example via a sum
\begin{equation}\label{convlayer}
\sum_g |\scal{x}{gt}+b|_+,  \quad t\in \TT, b\in \R\
\end{equation}
or a max operation

\begin{equation}\label{convlayer2}
\max_g |\scal{x}{gt}+b|_+,  \quad t\in \TT, b\in \R\
\end{equation}

This  corresponds to the neural response of a so called complex cell in \cite{Hubel1962}.

%%%%%%%%%%%%%%%%%%
\subsection{A HW module is a kernel machine}

It is trivial that almost any (positive or negative defined) nonlinearity after the dot product
in a network yields a kernel. Consider a 3-layers network with the first layer
being the input layer ${x}$. Unit $i$ in the second layer (comprising
$N$ units) computes $|\scal{t_i}{x} + b_i|_+ = \phi_i(x),
i=1,\cdots,N$. Thus each unit in the third layer performing a dot
product of the vector of activities from the second layer with weights
$\phi_i(y)$ (from layer two to three) computes $K(x,y) = \sum_j
\phi_j(x)\phi_j(y)$ which is guaranteed to be well defined because the
sum is finite. Key steps of the formal proof, which holds also in the
general case,  can be found in Rosasco and Poggio,
\cite{Rosascokernels2015} and references there; see also Appendix \ref{summaryISK} and
Appendix \ref{ExampleKernel} for an example.\\

\begin{figure}
\includegraphics[width=0.6\textwidth]{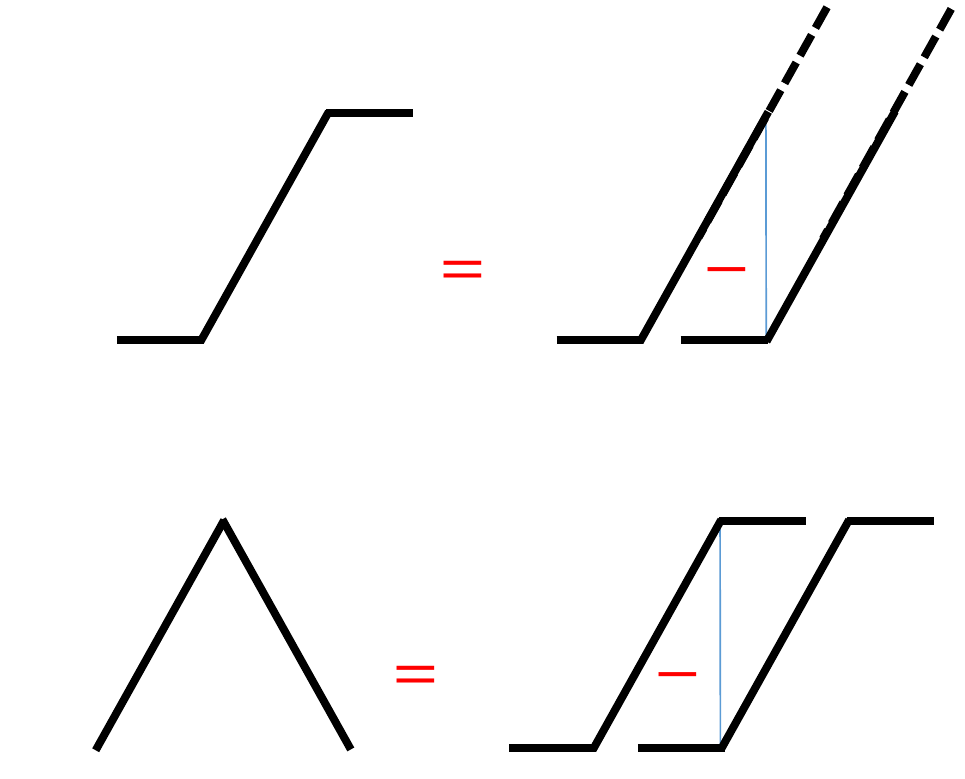}
\caption{{\it A ``Gaussian'' of one variable can be written as the linear
  combination of ramps (e.g. rectifiers): a) a sigmoid-like function can be written as linear combinations of ramps b)
  linear combinations of sigmoids give gaussian-like triangular functions.}\label{equivalence}}
\end{figure}

A different argument shows that it is ``easy'' to obtain kernels in
layered networks. Assume that inputs $x \in \R^n$ as well as the
weights $t$ are normalized, that is $x,t \in S^n$ where $S^n$ is the unit sphere.\\
\noindent
In this case dot products are radial functions (since  $\scal{x}{t} =
\frac{1}{2} (2 -(|x-t|^2))$) of $r^2$. The kernel $r^2$ can be shaped
  by linear combinations (with bias) of rectifier units (linear
  combinations of ramps can generate sigmoidal units, linear
  combinations of which are equivalent to quasi-Gaussian,
  ``triangular'' functions) \footnote{Notice that in one dimension the
    kernel $|x-y|$ can be written in terms of ramp functions as $|x-y|
    =|x-y|_+ + |-(x-y)|_+$. See
  Figure \ref{equivalence} and \cite{GirJonPog95}.}

Thus for normalized inputs dot products with nonlinearities can easily be
equivalent to radial kernels. Pooling before a similarity
operation (thus pooling at layer $n$ in a DCN before dot products in
layer $n+1$) maintains the kernel structure (since $\widetilde K
(x,x')=\int dg \int dg' K(gx,g'x')$ is a kernel if $K$ is a
kernel). Thus
{\it DCNs with normalized inputs are hierarchies of radial kernel machines,
 also called Radial Basis Functions (RBFs)}.

Kernels induced by  linear combinations of features such as

\begin{equation}\label{convlayer}
\sum_g |\scal{x}{gt}+b|_+,  \quad t\in \TT, b\in \R\
\end{equation}

\noindent are selective and invariant (if $G$ is compact, see Appendix
\ref{summaryISK}). The max operation $\max_g |\scal{x}{gt}+b|_+$ has a
different form and does not satisfy the condition of the theorems in
the Appendix. On the other hand the pooling defined in terms of the
following soft-max operation (which approximates the max for ``large'' $n$)

\begin{equation}\label{softmax2}
\sum_g \frac{(\scal{x}{gt})^n}{\sum_{g'} (1+\scal{x}{g't})^{n-1}},  \quad t\in \TT\,
\end{equation}

\noindent induces a kernel that satisfies the conditions of the
theorems summarized  in Appendix \ref{summaryISK}. It is well known that such an
operation can be implemented by simple circuits of the lateral
inhibition type (see \cite{kouh2008canonical}). On the other hand the
pooling in Appendix \ref{Mex} (see \cite{CohenS14}) does not correspond in general to a
kernel, does not satisfy the conditions of Appendix \ref{summaryISK}
and is not guaranteed to be selective.

Since weights and ``centers'' of the RBFs are learned in a supervised way, the kernel
machines should be more properly called HyperBF, see Appendix
\ref{HBF}.

%%%%%%%%%%%%%%%%%%%%
\subsection{Summary: every layer of a DCN is  a kernel machine}

Layers of a Deep Convolutional Network using linear rectifiers
(ramps) can be described as

\begin{equation}\label{rectlayer}
\sum_g |\scal{x}{gt}+b|_+,  \quad t\in \TT, b\in \R\
\end{equation}

\noindent where the range of pooling ($\sum_g$) may be degenerate (no
pooling) in which case the layer is not a convolutional layer.

Such a layer corresponds to the kernel

\begin{equation}\label{kernel}
\tilde{K}(x,x')=\int\;dg\;dg'\;K_{0}(x,g,g').
\end{equation}

\noindent with

\begin{equation}\label{kernel}
K_{0}(x,g,g')=\int\;dt\;db\;|\scal{gt}{x}+b|_{+}|\scal{g't}{x'}+b|_{+}.
\end{equation}

An alternative pooling to \ref{rectlayer} is provided by softmax pooling

\begin{equation}\label{softlayer}
\sum_g \frac{(\scal{x}{gt})^n}{\sum_{g'} (1+\scal{x}{g't})^{n-1}},  \quad t\in \TT\,
\end{equation}

Present-day DCNs seem to use equation \ref{softlayer} for the pooling
layers and equation \ref{rectlayer} for the non-convolutional
layers. The latter is the degenerate case of equation \ref{softlayer}
(when $G$ contains only the identity element).

%%%%%%%%%%%%%%%%%%%%

\section{A conjecture on what a hierarchy does}

The previous sections describe an extension of i-theory that can be
applied exactly to any layer of a DLCN and any of the nonlinearities
that have been used: pooling, linear rectifier, sigmoidal
units. In this section we suggest a framework for understanding
hierarchies of such modules, which we hope may lead to new formal or
empirical results.

\begin{figure}\centering
\includegraphics[width=0.7\textwidth]{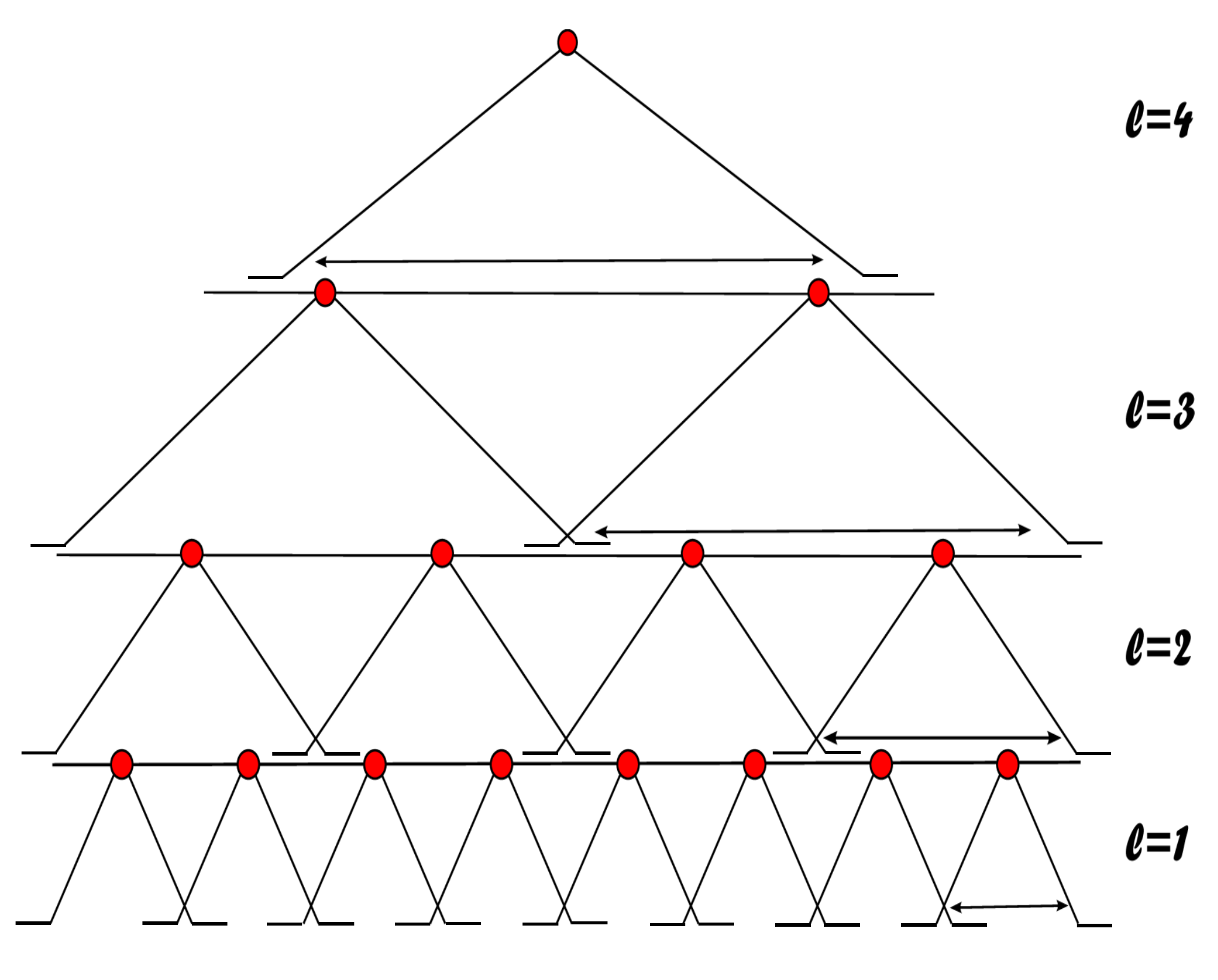}
\caption{{\it A hierarchical, supervised architecture built from eHW-modules. Each red
  circle represents the signature vector computed by the associated
  module (the outputs of complex cells) and double arrows represent
  its receptive fields -- the part of the (neural) image visible to
  the module (for translations this is also the pooling range). The
  ``image'' is at level $0$, at the bottom.}\label{Figure1}}
\end{figure}

\subsection{DLCNs and HVQ}\label{DLCNsHVQ}

We consider here an HW module with pooling (no pooling corresponds to 1
pixel stride convolutional layers in DLN). Under the assumption of
normalized inputs, this is equivalent to a HBF module with Gaussian-like radial
kernel and  ``movable'' centers (we assume standard Euclidean distance,
 see Poggio, Girosi,\cite{Poggio1990b}; Jones, Girosi and Poggio,\cite{GirJonPog95}).

Notice that one-hidden-layer HBF can be much more efficient in storage
(e.g. bits used for all the centers) than classical RBF because of the
smaller number of centers (HBFs are similar to a multidimensional
free-knots spline whereas RBFs correspond to classical spline).

\begin{figure}
\includegraphics[width=1.1\textwidth]{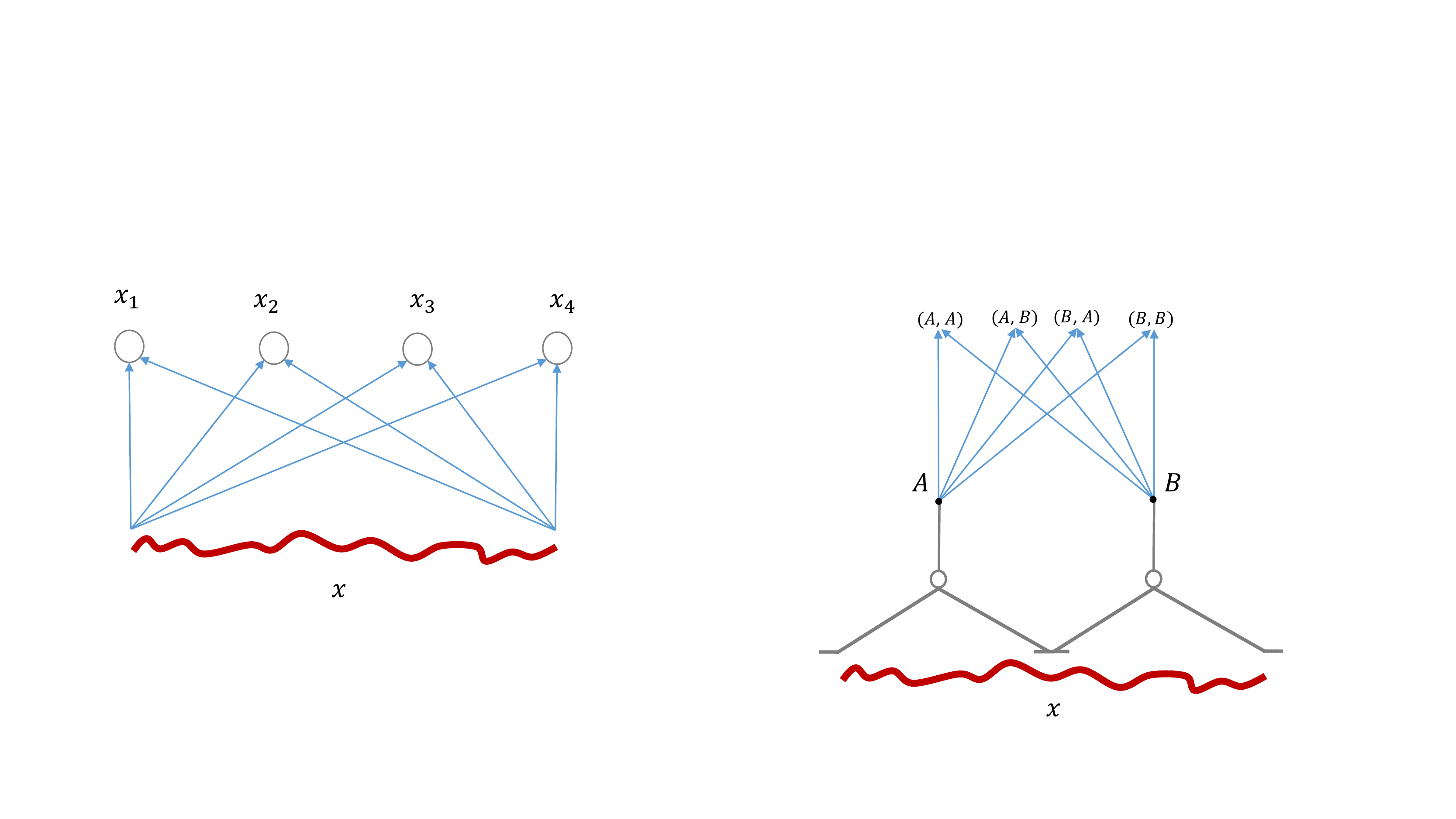}
\caption{{\it See text, Example in section \ref{DLCNsHVQ}}.\label{Example}}
\end{figure}

The next step in the argument is the observation that a network of
radial Gaussian-like units become in the limit of $\sigma \to 0$ a
look-up table with entries corresponding to the centers. The network
can be described in terms of {\it soft Vector Quantization} (VQ) (see
section 6.3 in Poggio and Girosi, \cite{Poggio89atheory}). Notice that
hierarchical VQ (dubbed HVQ) can be even more efficient than VQ in
terms of storage requirements (see e.g. \cite{HVQ}). This suggests
that a hierarchy of HBF layers may be similar (depending on which
weights are determined by learning) to HVQ. Note that {\it compression is
  achieved when parts can be reused in higher level layers as in
  convolutional networks}. Notice
that the center of one unit at level $n$ of the ``convolutional'' hierarchy of Figure
\ref{Figure1}  is a combinations of parts provided by each of the
lower units feeding in it. This may  even happen without convolution
and pooling as shown
in the following extreme example.\\\\
\noindent
{\bf Example}
\noindent
Consider the case of kernels that are in the limit delta-like
functions (such as Gaussian with very small variance). Suppose as in
Figure \ref{Example} that there are four possible quantizations of the
input $x$: $x_1, x_2, x_3, x_4$. One hidden layer would consist of
four units $\delta(x-x_i), i=1,\cdots,4$. But suppose that the
vectors $x_1, x_2, x_3,x_4$ can be decomposed in terms of two smaller
parts or features $x'$ and $x"$, e.g.  $x_1 =x'\oplus x"$,
$x_2=x'\oplus x'$, $x_3=x"\oplus x"$ and $x_4=x"\oplus x'$. Then a two
layer network could have two types of units in the first layer
$\delta(x-x')$ and $\delta(x-x")$; in the second layer four units will
detect the conjunctions of $x'$ and $x"$ corresponding to
$x_1, x_2, x_3,x_4$. The memory requirements will go from $4N$ to
$2N/2+8$ where $N$ is the length of the quantized vectors; the latter
is much smaller for large $N$. Memory compression for HVQ vs VQ --
that is for multilayer networks vs one-layer networks -- increases
with the number of (reusable) parts.  Thus for problems that are {\it
  compositional}, such as text and images, hierarchical
architectures of HBF modules minimize memory requirements.

Classical theorems (see refrences in \cite{GirPog-Kol89,GirPog-best90} show that
one hidden layer networks
can approximate arbitrarily well rather general classes of
functions. A possible advantage of multilayer vs one-layer
networks that emerges from the analysis of this paper is memory
efficiency which can be critical for large data sets and is related to
generalization rates.

\section{Remarks}

\begin{itemize}

\item Throughout this note, we discuss the potential properties of
  multilayer networks, that is the properties they have with the
  ``appropriate'' sets of weights when supervised training is
  involved. The assumption is therefore that greedy SGD using very
  large sets of labeled data, can find the ``appropriate sets of sets
  of weights".

\item Recently, several authors have expressed surprise when observing
  that the last hidden unit layer contains information about tasks
  different from the training one (e.g. \cite{Yamins2014}). This
  is in fact to be expected. The last layer of HBF is rather
  independent of the training target and mostly depends on the input
  part of the training set (see theory and gradient descent equations
  in Poggio and Girosi, \cite{Poggio89atheory} for the one-hidden layer case). This is
  exactly true for one-hidden layer RBF networks and holds
  approximatively for HBFs. The weights from the last hidden layer to
  the output are instead task/target dependent.
\item The result that linear combinations of rectifiers can be
  equivalent to kernels is {\it robust} in the sense that it is true
  for several different nonlinearities such as rectifiers, sigmoids
  etc. Ramps (e.g. rectifiers) are the most basic ones.  {\it Such
    robustness is especially attractive for neuroscience.}

\end{itemize}

\newpage
%%%%%%%%%%%%%%%%%%%%%%%%%%%%%%%%%%%%%%%%%%%%%%%

\section{Appendices}

\subsection{HW modules are equivalent to kernel machines (a summary of the results in \cite{Rosascokernels2015})}\label{summaryISK}
In the following we summarize the key passages of
\cite{Rosascokernels2015} in proving that HW modules are kernel machines:
\begin{enumerate}
\item
The feature map
$$
\phi(x,t,b) = |\scal{t}{x}+b|_{+}
$$
(that can be associated to the output of a simple cell, or the basic computational unit of a deep learning architecture)
can also be seen as a kernel in itself. The kernel can be a universal kernel. In fact, under the hypothesis of normalization of the vectors $x,t$ we have that $2(|1-\scal{t}{x}|_{+}+|\scal{t}{x}-1|_{+})=2|1-\scal{t}{x}|=\nor{x-t}_{2}^{2}$ which is a universal kernel (see also th 17 of \cite{micchelli2006}).\\
%Let $\tilde{x}=(x,1)^{T},\;\tilde{t}=(t,b)^{T}$. Using the
%result in Appendix \label{triang}, in the hypothesis that $\nor{\tilde{x}}=\nor{\tilde{t}}=1$, we have that
%$$
%\phi(x,t,b)=|\scal{\tilde{t}}{\tilde{x}}|_{+}\sim e^{-\nor{\tilde{x}-\tilde{t}}^{2}}
%$$
\noindent
The feature $\phi$ leads to a kernel
$$
K_{0}(x,x') = \phi^{T}(x)\phi(x') = \int\;db\;dt\;|\scal{t}{x}+b|_{+}|\scal{t}{x'}+b|_{+}
$$
which is a universal kernel being a kernel mean embedding (w.r.t. $t,b$, see \cite{SrGrFu10}) of the a product of universal kernels.
\item
If we explicitly introduce a group of transformations acting on the feature map input i.e.
$\phi(x,g,t,b) = |\scal{gt}{x}+b|_{+}$ the associated kernel can be written as
$$
\tilde{K}(x,x')=\int\;dg\;dg'\int\;dt\;db\;|\scal{gt}{x}+b|_{+}|\scal{g't}{x'}+b|_{+}=\int\;dg\;dg'\;K_{0}(x,g,g').
$$
$\tilde{K}(x,x')$ is the group average of $K_{0}$ (see \cite{burkhardt2007}) and can be seen as the mean kernel embedding of $K_{0}$ (w.r.t. $g,g'$, see \cite{SrGrFu10}).
\item
The kernel $\tilde{K}$ is invariant and, if $G$ is compact, selective i.e.
$$
\tilde{K}(x,x')=1\;\Leftrightarrow\;x\sim x'.
$$
The invariance follows from the fact that any $G-$group average function is invariant to $G$ transformations. Selectivity follows from the fact that $\tilde{K}$ a universal kernel being a kernel mean embedding of $K_{0}$ which is a universal kernel (see \cite{SrGrFu10}).
\end{enumerate}
\begin{remark}
If the distribution of the templates $t$ follows a gaussian law the kernel $K_{0}$, with an opportune change of variable, can be seen as a particular case of the $n$th order arc-cosine kernel in \cite{cho2009} for $n=1$.
\end{remark}

\subsection{An example of an explicit calculation for an inner product kernel}\label{ExampleKernel}
Here we note how a simple similarity measure between functions of the form in eq. \eqref{convlayer} correspond to a kernel in the case when the inner product between two HW module outputs at the first layer, say $\mu(I),\mu(I')$, is calculated using a step function nonlinearity.
Note first that a heaviside step function can be approximated by a sigmoid like function derived by a linear combination of rectifiers of the form:
$$
H(x)\sim \alpha(|x|_{+}-|x-\frac{1}{\alpha}|_{+})
$$
for very large values of $\alpha$.
%Since each CDF is square integrable (it follows from the fact that $|\scal{f_\rho}{f_{\rho'}}|\leq \nor{f_{\rho}}|_{\infty}\nor{f_{\rho'}}|_{1}$), we can endow the CDF space ${\mathcal F}(\R)$ with the usual $L^2(\R)$ inner product and note that
With this specific choice of the nonlinearity  we have that the inner product of the HW modules outputs (for fixed $t$) at the first layer is given by:
$$
\scal{\mu(I)}{\mu(I')} = \int \int dg\;dg'( \int db H(b-\scal{I}{gt})H(b-\scal{I'}{g't})).
$$
with
$$
\mu^{t}_{b}(I)=\int\;dg\;H(b-\scal{I}{gt}).
$$
Assuming that the scalar products, $\scal{I}{gt},\scal{I'}{g't}$, range in the interval $[-p,p]$ a direct computation of the integral above by parts shows that, $x=\scal{I}{gt}, x'=\scal{I'}{gt}$:
\begin{align*}
& \int_{-p}^{p} db H(b-x)H(b-x')\\
&= H(b-x)\big((b-x')H(b-x')-(x-x')H(x-x')\big)\rvert_{-p}^{p}\\
&= p-\frac{1}{2}(x+x'+|x-x'|)
\end{align*}
and being
\begin{align*}
\max\{x, x'\} &= \max\left\{x - \frac{1}{2}(x+x'), x' - \frac{1}{2}(x+x')\right\} + \frac{1}{2}(x+x')\\
&= \max\left\{\frac{1}{2}(x'-x), \frac{1}{2}(x - x')\right\} + \frac{1}{2}(x+x')\\
&= \max\left\{-\frac{1}{2}(x-x'), \frac{1}{2}(x-x')\right\} + \frac{1}{2}(x+x')\\
&= +\left|\frac{1}{2}(x-x')\right| + \frac{1}{2}(x+x')\\
&= \frac{1}{2}(x+x'+|x-x'|).
\end{align*}
we  showed that
$$
K(\scal{I'}{t}, \scal{I}{t}) = C-\max(\scal{I'}{t},\scal{I}{t})
$$
which defines a kernel. If we include the pooling over the transformations and templates we have
\begin{equation}\label{kernelembedding}
\tilde{K}(I,I') = p-\int\;d\lambda(t)\;dg\;dg'\;\max(\scal{I'}{gt},\scal{I}{g't}).
\end{equation}
where $\lambda(t)$ is a probability measure on the templates $t$. $\tilde{K}$ is again a kernel.\\
A similar  calculation can be repeated at the successive layer leading to kernels of kernels structure.

\subsection{Mex}
\label{Mex}
Mex is a generalization of the pooling function. From \cite{CohenS14} eq. 1 it is defined as:
\begin{equation}\label{Mexeq}
Mex_{(\{c_{i}\},\xi)}=\frac{1}{\xi}\log\Big(\frac{1}{n}\sum_{i=1}^{n}\exp(\xi c_{i})\Big)
\end{equation}
We have
\begin{eqnarray*}
&& Mex_{(\{c_{i}\},\xi)}\xrightarrow[\xi \to \infty]{} Max_{i}{(c_{i})}\\
&& Mex_{(\{c_{i}\},\xi)}\xrightarrow[\xi \to 0]{} Mean_{i}{(c_{i})}\\
&& Mex_{(\{c_{i}\},\xi)}\xrightarrow[\xi \to -\infty]{} Min_{i}{(c_{i})}.
\end{eqnarray*}
We can also choose values of $\xi$ in between the ones above, the interpretation is less obvious.
The Mex pooling does not define a kernel since is not positive definite in general (see also Th 1 in \cite{CohenS14})

\subsection{Hyper Basis Functions: minimizing memory in Radial Basis Function networks}
\label{HBF}

We summarize here an old extension by Poggio and Girosi  \cite{PoGir94extension} of the
classical kernel networks called Radial Basis Functions (RBF). In summary (but see the paper) they
extended the theory by defining a general form of these networks which
they call Hyper Basis Functions.  They have two sets of modifiable
parameters: {\it moving
  centers} and {\it adjustable norm-weights}.  Moving the centers is
equivalent to task-dependent clustering and changing the norm weights
is equivalent to task-dependent dimensionality reduction.\\
\noindent
A classical RBF has the form
\begin{equation}\label{eq:green-rad}
f({\bf x}) = \sum_{i = 1}^N  c_i  K(\| {\bf x} -
{\bf x}_i \|^2),
\end{equation}
\noindent which is a sum of radial functions, each with its {\it
center} ${\bf x}_i$ on a distinct data point. Thus the
number of radial functions, and corresponding centers, is
the same as the number of examples. Eq. \eqref{eq:green-rad} is a minimizer solution of
\begin{equation}\label{H}
H[f] = \sum_{i=1}^{N} (y_{i}-f(x_{i}))^{2} + \lambda\nor{Pf}^{2}\;\;\lambda\in \R^{+}
\end{equation}
where $P$ is a constrain operator (usually a differential operator).\\
\noindent
HBF extend RBF in two directions:
\begin{enumerate}
\item The computation of a solution of the form \eqref{eq:green-rad}
  has a complexity (number of radial functions) that is independent of
  the dimensionality of the input space but is on the order of the
  dimensionality of the training set (number of examples), which can
  be very high.  Poggio and Girosi showed how to justify an
  approximation of equation \eqref{eq:green-rad} in which the number
  of centers is much smaller than the number of examples and the
  positions of the centers are modified during learning. The key idea is to consider a specific form of an
  approximation to the solution of the standard regularization
  problem.
\item Moving centers are equivalent to the free knots of
 nonlinear splines. In the context of networks they were first
  suggested as a potentially useful heuristics by Broomhead and Lowe
\cite{BroLow88} and used by Moody and Darken   \cite{MooDar89}.
\end{enumerate}
\noindent
Poggio and Girosi called {\it Hyper Basis Functions}, in short {\it HyperBFs}, the most
general form of regularization networks based on these
extensions plus the use of a weighted norm.

\subsubsection{Moving Centers}
The solution given by standard regularization theory to the
approximation problem can be very expensive in computational terms
when the number of examples is very high.  The computation of the
coefficients of the expansion can become then a very time consuming
operation: its complexity grows polynomially with $N$, (roughly as
$N^3$) since an $N\times N$ matrix has to be inverted. In addition,
the probability of ill-conditioning is higher for larger and larger
matrices (it grows like $N^3$ for a $N \times N$ uniformly distributed
random matrix) \cite{Demmel87}. The way suggested by
Poggio and Girosi to reduce the complexity of the problem is as follows. While the exact
regularization solution is equivalent to generalized splines with {\it
  fixed} knots, the approximated solution is equivalent to generalized
splines with {\it free} knots.

A standard technique, sometimes known as Galerkin's method, that has
been used to find approximate solutions of variational problems, is to
expand the solution on a finite basis.  The approximated solution
$f^*({\bf x})$ has then the following form:
\begin{equation}
f^*({\bf x}) = \sum_{i = 1}^n c_{i} \phi_{i}({\bf x})
\label{eq:galerkin}
\end{equation}
\noindent where $\{ \phi_{i} \}_{i = 1}^n$ is a set of
linearly independent functions \cite{Mikhlin65}. The
coefficients $c_{i}$ are usually found according to some rule that
guarantees a minimum deviation from the true solution. A natural approximation to the exact solution
will be then of the form:

\begin{equation}
f^*({\bf x}) = \sum_{\alpha = 1}^n c_{\alpha} G({\bf x} ; {\bf
t}_{\alpha})
\label{eq:GRBF}
\end{equation}

\noindent where the parameters ${\bf t}_{\alpha}$, that we call
``centers'', and the coefficients $c_{\alpha}$ are unknown, and are in
general fewer than the data points ($n \leq N$).  This form of
solution has the desirable property of being an universal approximator
for continuous functions \cite{Poggio89atheory} and to be the only choice that guarantees that
in the case of $n = N$ and $\{ {\bf t}_{\alpha}\}_{\alpha = 1}^n = \{
{\bf x}_i\}_{i = 1}^n$ the correct solution (of equation
\eqref{H} ) is consistently recovered. We will see later how to
find the unknown parameters of this expansion.

\subsubsection{How to learn centers' positions}

Suppose that we look for an approximated solution of the
regularization problem of the form
\begin{equation}\label{eq:GRBF-W}
f^*({\bf x}) = \sum_{\alpha = 1}^n c_{\alpha} G(\nor{{\bf x}- {\bf
t}_{\alpha}}^{2})
\end{equation}
We now have
the problem of finding the $n$ coefficients $c_{\alpha}$, the $d
\times ~n$ coordinates of the centers ${\bf t}_{\alpha}$. We can
use the natural definition of optimality given by the functional $H$.
We then impose the condition that the set $\{ c_{\alpha}, {\bf
t}_{\alpha} | \alpha = 1, ..., n \}$  must be such that they
minimizes $H[f^*]$, and the following equations must be satisfied:
$$
{\partial H[f^*] \over \partial c_{\alpha}} = 0~,~~~~~
{\partial H[f^*] \over \partial {\bf t}_{\alpha}} = 0, ~~~
\alpha = 1, ..., n~.
$$
Gradient-descent is probably the simplest approach for attempting to
find the solution to this problem, though, of course, it is not
guaranteed to converge.  Several other iterative methods, such as
versions of conjugate gradient and simulated annealing
\cite{KirGelVec83} may be more efficient than gradient
descent and should be used in practice. Since the function $H[f^*]$ to
minimize is in general non-convex, a stochastic term in the gradient
descent equations may be advisable to avoid local minima.  In the
stochastic gradient descent method the values of $c_{\alpha}$, ${\bf
t}_{\alpha}$ and $\bf M$ that minimize $H[f^*]$ are regarded as the
coordinates of the stable fixed point of the following stochastic
dynamical system:
$$
\dot c_\alpha =  - \omega
{\partial H[f^*] \over \partial c_\alpha} + \eta_\alpha (t),  ~~\alpha = 1, \dots , n
$$
$$
\dot {\bf t}_\alpha =  - \omega
{\partial H[f^*] \over \partial {\bf t}_\alpha} + \mbox{\boldmath $\mu$}_\alpha (t),  ~~\alpha = 1, \dots , n
$$
\noindent where $\eta_\alpha (t)$, $\mbox{\boldmath $\mu$}_\alpha (t)$
are white noise of zero mean and $\omega$ is a parameter determining
the microscopic timescale of the problem and is related to the rate of
convergence to the fixed point. Defining
$$
\Delta_i \equiv  y_i - f^*({\bf x}) =
y_i - \sum_{\alpha = 1}^n c_\alpha G(\| {\bf x}_i -
{\bf t}_\alpha \|^2)
$$
\noindent we obtain
$$
H[f^*] = H_{{\bf c}, {\bf t}}= \sum_{i=1}^{N} (\Delta_i)^2.
$$
The important quantities -- that can be used in more efficient schemes
than gradient descent -- are

\begin{itemize}
\item for the $c_\alpha$
\begin{equation}\label{eq:grad-c}
{{\partial H[f^*]} \over {\partial c_\alpha}} = - 2 \sum_{i = 1}^N
\Delta_i G(\| {\bf x}_i - {\bf t}_\alpha \|^2)~~;
\end{equation}
\item for the centers $t_\alpha$
\begin{equation}\label{eq:grad-t}
{{\partial H[f^*]} \over {\partial {\bf t}_\alpha}} = 4 c_\alpha
\sum_{i = 1}^N \Delta_i G'(\| {\bf x}_i - {\bf
t}_\alpha \|^2)({\bf x}_i - {\bf
t}_\alpha)
\end{equation}
\end{itemize}
\noindent
{\bf Remarks}
\begin{enumerate}
\item Equation \eqref{eq:grad-c} has a simple interpretation: the
  correction is equal to the sum over the examples of the products
  between the error on that example and the ``activity'' of the
  ``unit'' that represents with its center that example.  Notice that
  $H[f^*]$ is quadratic in the coefficients $c_\alpha$, and if the
  centers are kept fixed, it can be shown \cite{Poggio89atheory} that
  the optimal coefficients are given by
\begin{equation}\label{eq:pseudo}
{\bf c} = (G^T~G + \lambda g)^{-1} G^T {\bf y}
\end{equation}
\noindent where we have defined $({\bf y})_i = y_i$, $({\bf
c})_\alpha = c_\alpha$, $(G)_{i \alpha} = G( {\bf x}_i ; {\bf
t}_\alpha)$ and $(g)_{\alpha \beta} = G({\bf t}_\alpha ; {\bf
t}_{\beta})$.  If $\lambda$ is let go to zero, the matrix on the right
side of equation \eqref{eq:pseudo} converges to the pseudo-inverse of
$G$ \cite{Albert72} and if the Green's function is
radial the approximation method of  \cite{BroLow88} is recovered.
\item Equation \eqref{eq:grad-t} is similar to task-dependent
  clustering \cite{Poggio89atheory}. This can be best seen by assuming
  that $\Delta_i$ are constant: then the gradient descent updating
  rule makes the centers move as a function of the majority of the
  data, that is of the position of the clusters. In this case a
  technique similar to the k-means algorithm is recovered,
  \cite{MacQueen67,MooDar89}. Equating ${\partial H[f^*]} \over
  {\partial {\bf t}_\alpha}$ to zero we notice that the optimal
  centers ${\bf t}_\alpha$ satisfy the following set of nonlinear
  equations:
$$
{\bf t}_\alpha = {{\sum_i P_i^\alpha {\bf x}_i} \over
{\sum_i P_i^\alpha}} ~~~\alpha = 1, \dots , n
$$
\noindent where $P_i^\alpha = \Delta_i G'(\| {\bf x}_i -
{\bf t}_\alpha \|^2)$.  The optimal centers are then a weighted sum of
the data points. The weight $P_i^\alpha$ of the data point $i$ for a
given center ${\bf t}_\alpha$ is high if the interpolation error
$\Delta_i$ is high there {\em and} the radial basis function centered
on that knot changes quickly in a neighborhood of the data point.
This observation suggests faster update schemes, in which a suboptimal
position of the centers is first found and then the $c_\alpha$ are
determined, similarly to the algorithm developed and tested
successfully by Moody and Darken \cite{MooDar89}.
\end{enumerate}

\subsubsection{An algorithm}

It seems natural to try to find a reasonable initial value for the
parameters ${\bf c}, {\bf t}_\alpha$, to start the stochastic minimization
process. In the absence of more specific prior information the following
heuristics seems reasonable.
\begin{itemize}
\item Set the number of centers and set  the centers' positions
to positions suggested by cluster analysis of the data (or more simply
to a subset of the examples' positions).
\item Use matrix pseudo-inversion to find the $c_\alpha$.
\item Use the ${\bf t}_\alpha$, and $c_\alpha$ found so far as initial
  values for the stochastic gradient descent equations.
\end{itemize}
Experiments with movable centers and movable weights have been
performed in the context of object recognition (Poggio and Edelman, \cite{PogEde90}; Edelman and Poggio, \cite{EdePog90}) and
approximation of multivariate functions.

\subsubsection{Remarks}
\begin{enumerate}
\item Equation  \eqref{eq:grad-t} is similar to a
clustering process.
\item In the case of $N$ examples, $n=N$ fixed centers, there are
  enough data to constrain the $N$ coefficients $c_\alpha$ to be
  found.  Moving centers add another $n d$ parameters ($d$ is the
  number of input components). Thus the number of examples $N$ must be
  sufficiently large to constrain adequately the free parameters --
  $n$ d-dimensional centers, $n$ coefficients $c_\alpha$. Thus
$$
N>>n + nd.
$$
\end{enumerate}

\section*{Acknowledgment}

This work was supported by the Center for Brains, Minds and Machines
(CBMM), funded by NSF STC award  CCF – 1231216. This work was also supported by A*STAR JCO VIP grant $\#$1335h00098.
Part of the work was done in Singapore at the Institute for Infocomm
under REVIVE funding. TP thanks A*Star for its hospitality.

\bibliographystyle{ieeetr}
\bibliography{whydeepbib}

\end{document}